\documentclass[review, 11pt]{elsearticle}
\usepackage[numbers]{natbib}
\usepackage[margin=2.5cm]{geometry}
\journal{Applied Energy}

\usepackage[utf8]{inputenc}
\usepackage{graphicx}
\usepackage{booktabs}
\usepackage{changes}
\definechangesauthor[name={kevin}, color=orange]{km}
\definechangesauthor[name={tian}, color=blue]{tian}

\usepackage{hyperref}
\usepackage[capitalize, noabbrev, sort&compress]{cleveref}

\usepackage{ulem} 

\usepackage{caption}
\usepackage{subcaption}
\usepackage{comment}
\usepackage{float}
\usepackage{censor}
\usepackage{booktabs}
\usepackage{threeparttable}

\usepackage{tabularx}
\usepackage{multirow}

\usepackage{siunitx}
\usepackage{units}
\usepackage{amssymb}

\usepackage{setspace}
\usepackage{array}
\newcolumntype{L}[1]{>{\raggedright\let\newline\\\arraybackslash\hspace{0pt}}m{#1}}

\date{\today}

\usepackage{soul}

\begin{document}

\StopCensoring 

\begin{frontmatter}

\title{Estimating building energy efficiency from street view imagery, aerial imagery, and land surface temperature data}

\author[1]{Kevin Mayer \corref{cor1}}
\author[2]{Lukas Haas}
\author[1]{Tianyuan Huang}
\author[3]{Juan Bernabé-Moreno}
\author[1]{Ram Rajagopal}
\author[1]{Martin Fischer}

\address[1]{Department of Civil and Environmental Engineering, Stanford University, 473 Via Ortega, 94305, Stanford, USA}
\address[2]{Department of Computer Science, Stanford University, 353 Jane Stanford Way, 94305, Stanford, USA}
\address[3]{IBM Research Europe, IBM Europe, 200 Shelbourne Road, D04, Dublin, Ireland}
\cortext[cor1]{kdmayer@stanford.edu (corresponding author).}

\begin{abstract}
Current methods to determine the energy efficiency of buildings require on-site visits of certified energy auditors which makes the process slow, costly, and geographically incomplete. To accelerate the identification of promising retrofit targets on a large scale, we propose to estimate building energy efficiency from widely available and remotely sensed data sources only, namely street view, aerial view, footprint, and satellite-borne land surface temperature (LST) data. After collecting data for almost 40,000 buildings in the United Kingdom, we combine these data sources by training multiple end-to-end deep learning models with the objective to classify buildings as energy efficient (EU rating A-D) or inefficient (EU rating E-G). After evaluating the trained models quantitatively as well as qualitatively, we extend our analysis by studying the predictive power of each data source in an ablation study. We find that the end-to-end deep learning model trained on all four data sources achieves a macro-averaged F1 score of 64.64\% and outperforms the k-NN and SVM-based baseline models by 14.13 to 12.02 percentage points, respectively. Thus, this work shows the potential and complementary nature of remotely sensed data in predicting energy efficiency and opens up new opportunities for future work to integrate additional data sources. 
\end{abstract}


\end{frontmatter}

\section{Introduction}

Accounting for the whole building life cycle, the EU estimates that buildings are responsible for 40\% of the Union's energy consumption and for 36\% of its greenhouse gas emissions. While renovating the existing buildings could reduce the EU's total energy consumption and carbon emissions by approximately 5\%, less than 1\% of the national building stocks in the EU are, on average, renovated each year. This means that in order to meet its energy and climate objectives, the EU needs to at least double its current rate of renovations over the next years \cite{EuDirective}, requiring much more comprehensive and rapid evaluations of the existing building stock.

To accelerate the decarbonization of the existing building stock, the EU mandates building owners to provide building energy performance certificates whenever a building is sold or rented. In doing so, the EU aims to establish building energy efficiency as a price indicator to market players and hopes to incentivize owners of inefficient buildings to retrofit. While these certificates have enabled private buyers and tenants to compare a small number of properties more effectively, stakeholders in the retrofit sector still suffer from a lack of information as building energy performance databases are generally neither publicly available nor geographically complete. As a result, large-scale analyses to transform the existing building stock in a targeted way remain difficult. To facilitate this critical transformation of the building sector, we need technology-driven solutions to inform innovative policies and retrofit markets about individual buildings across geographies \cite{IEA2019, dupont}. When implemented, these solutions will enable a more efficient allocation of building retrofit investments, as market players can use building-level energy efficiency information to guide their actions.   

Based on the advent of large-scale, high-resolution public datasets in the remote sensing domain and recent advances in deep learning, this study explores the potential of artificial neural networks (ANN) to classify building energy efficiency from remotely sensed data sources only. Since this study relies on combining multiple high dimensional data sources in a non-linear way, we argue that training deep learning models in an end-to-end fashion should be the preferred modeling choice. By end-to-end deep learning models we mean ANN-based model architectures which learn to combine the different inputs non-linearly and without human supervision.

Our contribution to identify buildings with high retrofit potential on a large scale is three-fold. First, we create a dataset with more than 39,500 buildings by collecting building-level street view and aerial view imagery, footprint, and LST data. As such, this is the first study to collect and pre-process satellite-borne LST data to estimate building-level attributes such as building energy efficiency. Second, we train multiple end-to-end deep learning models to classify buildings as energy efficient (EU rating A-D) or inefficient (EU rating E-G) and analyze the resulting models quantitatively as well as qualitatively. Lastly, we extend our analysis by studying the predictive power of each data source in an ablation study and compare the performance of different end-to-end deep learning architectures. By ablation study we mean the procedure of removing certain input features or network elements in the end-to-end deep learning model in order to gain a better understanding of the respective contributions in terms of prediction performance. The model's predictions are compared against entries from the UK's official building energy performance registry \cite{EPC}.

\section{Related work}

Using machine learning-based methods to estimate building characteristics such as energy consumption \cite{Pham2020, Streltsov2020, Dougherty2021, Rosenfelder2021} and efficiency \cite{Kontokosta2012, Sun2022}, photovoltaic rooftop potential \cite{Lee2019, Krapf2021} and generation \cite{DSfG, Rausch2020, Mayer2022}, as well as property type, age, value, and footprint information \cite{Hoffmann2019, Bin2020, footprintextraction} has received significant research attention. In general, these studies can be further sub-divided into \textbf{top-down} approaches which start with estimates for a whole city or region and disaggregate them as needed and \textbf{bottom-up} approaches which in turn focus on individual buildings first \cite{Deb2021}. Since this paper estimates the energy efficiency for individual buildings, the subsequent review focuses on bottom-up approaches.
\subsection{Bottom-up approaches for building energy consumption and efficiency}

Predicting the energy consumption and efficiency of buildings is important because it can inform utility companies, residents, facility managers, contractors, and public agencies on how to improve the energy efficiency of the existing building stock \cite{berrill2022}. 

With the emergence of the first city-scale building-level benchmark datasets, earlier studies have focused on using tabular data to predict building energy consumption and efficiency. Incorporating information such as the building area, age, and the number of floors, \cite{Kontokosta2012} presents a regression-based approach for commercial buildings with more than 50,000 square feet. Similarly, \cite{Tsanas2012} develops a random forest model to predict energy-related building characteristics using tabular features such as a building's surface, wall, and roof area in order to estimate its heating load and cooling load.

In contrast to the studies relying on tabular data, \cite{Pham2020, Chou2018} utilize historical consumption data collected with smart meters in order to predict a building's short-term energy usage. While \cite{Pham2020} relies on a random forest-based model to estimate hourly building energy consumption, \cite{Chou2018} focuses on the comparison and evaluation of different modeling techniques, ranging from artificial neural networks, over support vector machines, to linear regression, and tree-based methods. 

Due to the lack of large-scale and publicly available building energy datasets, a growing body of research is studying building energy consumption and efficiency from remotely sensed data. Unlike previous approaches which are inherently limited to small geographic regions, the increasing availability of high-resolution remotely sensed data empowers this stream of research to potentially scale across geographies \cite{dupont}. In \cite{Streltsov2020}, the authors make use of overhead aerial imagery with a spatial resolution of 0.3m and publicly available building footprint information in order to derive estimates for residential energy consumption in a three-step procedure. After detecting and segmenting buildings in the overhead imagery, the authors classify the buildings by type into commercial and residential properties. The building energy consumption is then predicted with a random forest-based model which takes image-derived building features, i.e. footprint area and perimeter as well as the building type, as input. On a building-level, the model achieves an ${R}^2$ of 0.28 and 0.38 for the case studies in Gainesville, FL and San Diego, CA, respectively. \cite{Dougherty2021} extends this line of work in two ways. First, their work collects and analyzes two overhead images at different zoom-levels per building in order to better understand a building's spatial context. Second, the authors also generate a building-specific context vector based on the establishments within a given radius R. This fixed length context vector intends to capture the potential type of occupancy based on the social function of nearby establishments. Similarly, \cite{Rosenfelder2021} models building electricity consumption solely based on aerial and street view images. By adding street view images, the authors are able to achieve results which are comparable to conventional models based on public tabular datasets. As in \cite{Streltsov2020}, the authors find that spatially aggregating the predictions further improves the results.

Apart from the studies that focus on building energy consumption, \cite{Sun2022} presents a model which uses street view imagery and tabular data such as a building's total floor area, height, and number of open fireplaces in order to estimate a building's energy efficiency on a scale from A-G, a rating scheme introduced according to the EU's directive on the energy performance of buildings (EPBD), with "A" being the most energy efficient and "G" being the least efficient. In a case study for the city of Glasgow, more than 30,000 buildings are analyzed and the model achieves an accuracy of 86.8\%. 

While previous studies have used a combination of aerial and street view images to estimate building energy \textit{consumption}, methods to estimate building energy \textit{efficiency} from purely remotely sensed data have yet to be developed. Moreover, previous studies also do not include satellite-borne heat loss information derived from long wave infrared measurements. Hence, we extend the existing literature by predicting building energy efficiency from a new combination of purely remotely sensed data sources in an end-to-end deep learning model.

\section{Dataset}

\begin{table*}[ht]
    \small
    \caption{Dataset overview.}
    \label{tab:dataset-overview}
    \centering
    \begin{tabularx}{\textwidth}{lcccc}
    \hline
    Dataset split & Samples [\textit{Count}] & Samples [\%] & Geographies & Efficient buildings [\%] \\
    \hline
    Train & 32,315 & 81.59 & Coventry, Oxford, Westminster & 64.60 \\
    Validation & 3,590 & 9.06 & Coventry, Oxford, Westminster & 64.29 \\
    Test & 3,700 & 9.34 & Peterborough & 79.22 \\
    \hline
    Total & 39,605 & 100 & - & - \\
    \hline
    \end{tabularx}
\end{table*}

The dataset for our study consists of 39,605 buildings and spans multiple geographies as described in Table \ref{tab:dataset-overview}. Each building is represented by an aerial image, a street view image, satellite-borne heat loss measurements derived from LST data \cite{Gorelick2017, Ermida2020}, and OpenStreetMap-derived (OSM) footprint polygons \cite{OSM}. Moreover, each building has an associated ground truth energy performance label which specifies a building's energy efficiency in terms of seven classes ranging from "A", the best, to "G", the worst. The ground truth energy efficiency labels have been obtained from \cite{EPC}.

The ensure that our dataset represents the real world as closely as possible, we have collected building-level observations from four cities and regions across the United Kingdom: 21,607 buildings from the city of Coventry, 6,834 buildings from Westminster in the city of London, 7,464 buildings from the city of Oxford, and 3,700 buildings from the city of Peterborough. These cities have been selected based on the availability of the aforementioned data sources and their variety in urban landscapes. In the subsequent experiments, all models have been trained on an identical but randomly sampled training dataset consisting of 32,315 buildings from the cities of Coventry, Westminster, and Oxford, and a validation set of 3,590 buildings consisting of the remaining buildings from the aforementioned regions. The test set consists of 3,700 buildings and is exclusively taken from the city of Peterborough in order to better estimate the generalization performance of our models across unseen geographies.

It is important to note that the dataset exhibits a strong class imbalance. In the original dataset each building is assigned an energy performance label between "A" and "G", with 0.07\% of the buildings belonging to class \textit{A}, 2.68\% to \textit{B}, 16.83\% to \textit{C}, 46.36\% to \textit{D}, 26.29\% to \textit{E}, 5.85\% to \textit{F}, and 1.93\% to class \textit{G}. 
In this study, we aim to model the energy efficiency of a given building in a binary fashion, i.e. differentiate between efficient and inefficient buildings. To do so, the buildings in our dataset are grouped according to their respective energy performance label. Based on the input from an industry expert, buildings in the categories "A" to "D" are considered to be energy efficient (65.94\% of the data), while buildings in the categories "E" to "G" are considered to be inefficient (34.06\% of the data). The binary grouping significantly simplifies the modeling process, adapts better to data-scarce scenarios, and reduces the class imbalance while providing meaningful insights into the existing building stock.

To obtain the final dataset, the addresses in \cite{EPC} are geocoded and spatially joined with the OSM-based building footprints and the respective LST-derived heat loss signal. In cases where a given building consists of multiple flats and floors, only the label of the least efficient top floor apartment is considered for the sake of our estimation. This is because the satellite-borne heat loss measurements derived from \cite{Ermida2020} mostly capture a building's heat loss through its roof.

\subsection{Aerial and street view imagery}

The aerial and street view images are obtained from \cite{GoogleCloud}. For each building footprint, the coordinates of the respective centroid define the location for which we download the imagery. The street view images are downloaded with a field of view equal to 50 and the aerial images with a zoom level of 20.

\subsection{Land Surface Temperature (LST) data}

The land surface temperature data which provides the heat loss information is obtained from \cite{Gorelick2017} and builds upon \cite{Ermida2020} to obtain the Landsat-8-based long wave infrared product in an upsampled spatial resolution of 30x30m. The heat signal per building footprint is an average of all the building-specific LST observations for which the ground temperature at the time of data collection has been below a threshold of 5°C. The temperature threshold of 5°C has been determined in consultation with industry experts in the thermal imaging domain. 

\section{Methodology} 
\label{sec:methodology}

While previous studies have estimated building energy consumption from street view and aerial imagery \cite{Rosenfelder2021} or aerial imagery and footprint information \cite{Streltsov2020}, we shift the focus to predicting building energy \textit{efficiency}, introduce building-level heat loss information derived from LST observations as an entirely new data source, and fuse all these data sources in an end-to-end deep learning architecture as depicted in Figure \ref{fig:model_visualization}. 

\subsection{Data cleaning with K-Means clustering}

\begin{figure*}[h!]
    \centering
    \begin{minipage}[b]{1.0\linewidth}
    \centering
    \begin{subfigure}[b]{0.45\textwidth}
        \centering
        \includegraphics[width=\textwidth]{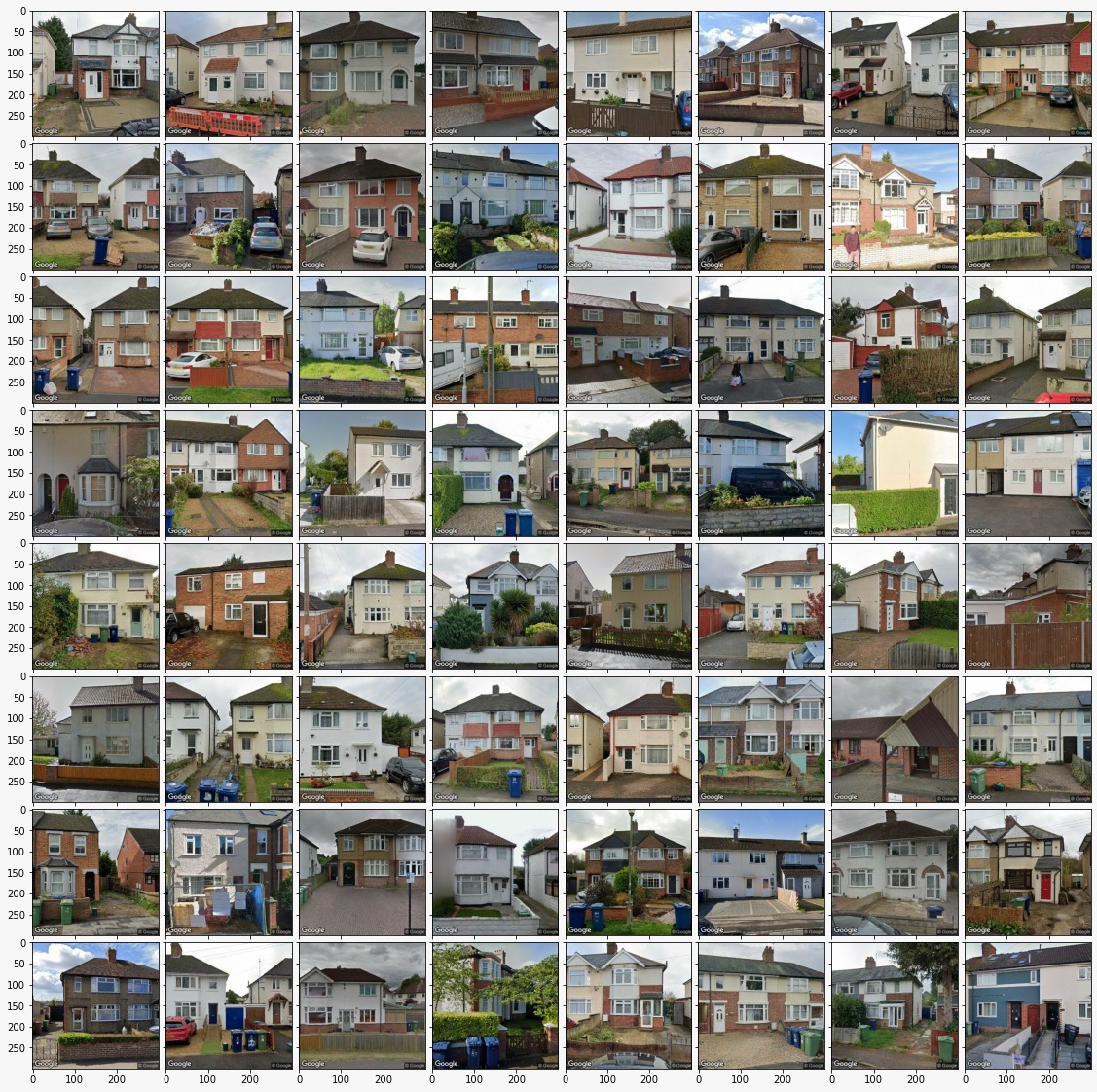}
        \caption{Clean street views}
    \end{subfigure}
    \hspace{8pt}
    \begin{subfigure}[b]{0.45\textwidth}
        \centering
        \includegraphics[width=\textwidth]{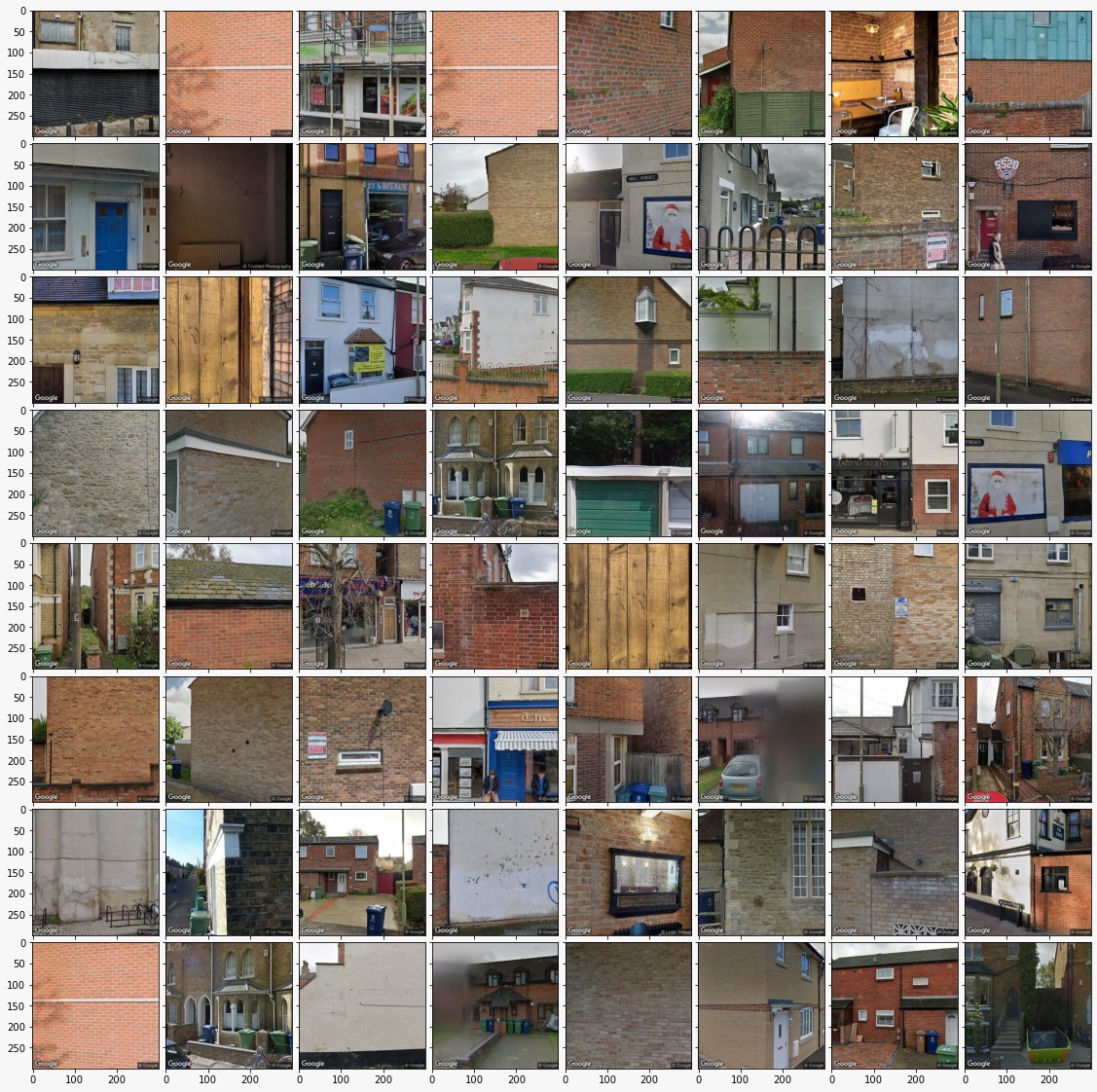}
        \caption{Noisy street views}
    \end{subfigure}
    \caption{Clusters representing clean and noisy street views.}
    \vspace{4pt}
    \label{fig:clusters}
    \end{minipage}
\end{figure*}

When working with multiple different data sources, ensuring a high dataset quality is essential to train effective models. This is particularly true for street view imagery, as this data source requires significant data cleaning efforts in order to remove noisy examples. In contrast, LST data obtained from \cite{Ermida2020} and aerial images obtained from \cite{GoogleCloud} have already been pre-processed and require minimal cleaning efforts only. When processing aerial images, we detect and remove empty or non-existent, i.e. "Sorry, we have no imagery here", images automatically based on a characteristic pixel combination in the respective server response. While experimenting with the LST data, we found that averaging the LST signal per building footprint across time achieves better results than calculating the median.

Cleaning the street view images is a multi-step process. First, all street view images are encoded into an embedding space with an Inception-v3-based encoder network \cite{inception}. Then, we conduct K-Means clustering on the embedded street view images in order to find images which do not depict meaningful features for our modeling task, i.e images which have been taken indoors or which do not show a building facade. To increase the quality of our dataset, image clusters without meaningful information are subsequently removed from the dataset. To verify the cluster-based decisions, we manually re-evaluate all removed street view images and add images which have been removed erroneously back to the cleaned dataset.

\begin{figure}[h]
    \centering
    \begin{minipage}[b]{1.0\linewidth}
    \centering
    \begin{subfigure}[b]{0.45\textwidth}
        \centering
        \includegraphics[width=\textwidth]{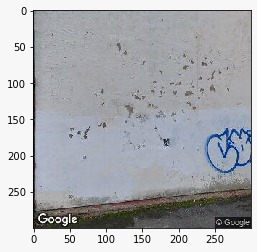}
        \caption{Reference image}
    \end{subfigure}
    \hspace{8pt}
    \begin{subfigure}[b]{0.45\textwidth}
        \centering
        \includegraphics[width=\textwidth]{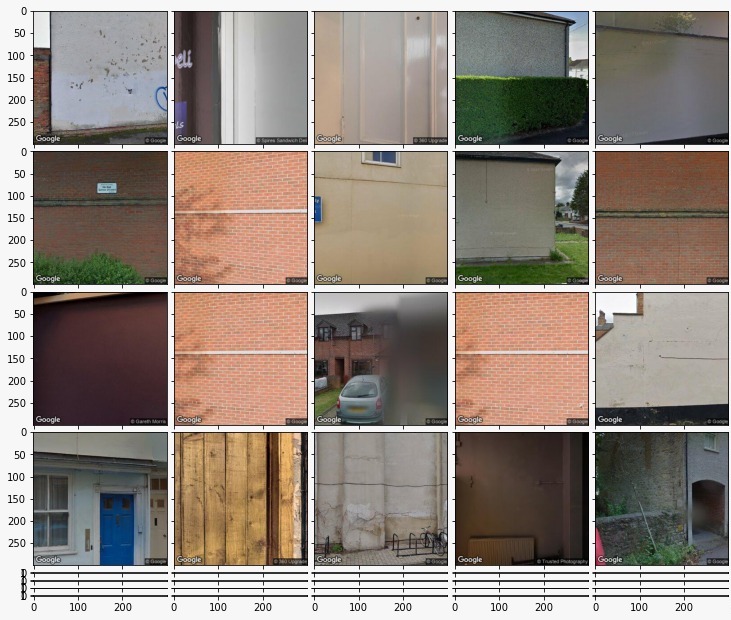}
        \caption{Nearest neighbors}
    \end{subfigure}
    \caption{A reference image and its nearest neighbors.}
    \vspace{4pt}
    \label{fig:nearest_neighbor}
    \end{minipage}
\end{figure}

Embedding and clustering the street view images with a pre-trained Inception-v3-based network reveals interesting patterns in our dataset. In Figure \ref{fig:clusters}, we juxtapose a cluster of street views depicting valid building images with a cluster of street views which mostly consists of brick walls and therefore bears little to no signal with respect to building energy performance. The embedded and clustered street views also provide insights into different urban environments and architectural styles which can be indicative of a property's socio-economic status \cite{socioeconomicsGSV} and construction date \cite{buildingageGSV}, as depicted in Figure \ref{fig:architectural_styles} in the appendix.

Based on the street view embeddings, we are also able to train a nearest neighbor model to perform a semantic search within the embedding space. This means that for a given image, we are able to automatically retrieve semantically similar images as illustrated in Figure \ref{fig:nearest_neighbor}. As a result, semantic search enables us to identify and analyze noisy street views in an automated fashion which significantly reduces the manual data cleaning effort.

\subsection{Data cleaning with semantic segmentation}
\label{data_cleaning_segmentation}

\begin{figure}[h!]
    \centering
    \begin{minipage}[b]{1.0\linewidth}
    \centering
    \begin{subfigure}[b]{0.45\textwidth}
        \centering
        \includegraphics[width=\textwidth]{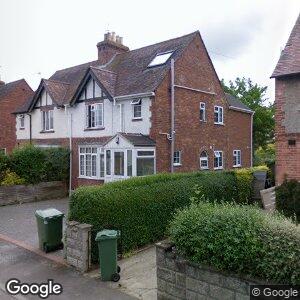}
        \caption{Original street view}
    \end{subfigure}
    \hspace{8pt}
    \begin{subfigure}[b]{0.45\textwidth}
        \centering
        \includegraphics[width=\textwidth]{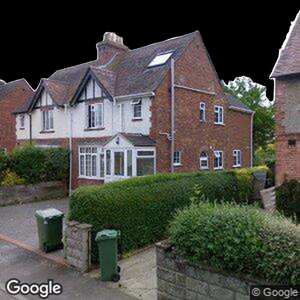}
        \caption{Segmented street view}
    \end{subfigure}
    \caption{Masking out the sky with semantic segmentation.}
    \vspace{4pt}
    \label{fig:street_view_seg}
    \end{minipage}
\end{figure}

Apart from cleaning the street view dataset with K-Means clustering and semantic search, we also use a pre-trained semantic segmentation model from \cite{inplace_abn}. Being trained on the \textit{Cityscapes} dataset \cite{Cityscapes} which has been developed for urban scene understanding, particularly autonomous driving, the segmentation model can be used as an optional pre-processing step in order to remove the sky from street view images, as illustrated in Figure \ref{fig:street_view_seg}. The intuition behind removing the sky from street view images is based on the fact that the sky does not provide any signal with respect to building energy performance but introduces a significant amount of noise through its variability in terms of location, date, and time of observation. Based on \cite{Li2015}, we decided to not mask out any other image features, such as cars, sidewalks, and vegetation, as these features can provide meaningful clues with respect to the spatial context and the socio-economic status of a given property \cite{gebru} and thus might correlate with building energy efficiency.

\subsection{Baselines}

For our baselines, we employ standard computer vision algorithms which work directly on the raw data. The baseline models include k-Nearest Neighbor (k-NN) and support vector machine (SVM) models and are trained for binary classification. Both baseline models are compared to a majority model (MM) which simply predicts the majority class for every data point. The rationale behind the choice of baselines is that both model types, SVMs and k-NNs, work well with high-dimensional data which is essential when working with image data input. To reduce the memory requirements and processing time, all our baseline models are trained on street view images only.

\subsection{End-to-end deep learning architecture}

\begin{figure*}[ht]
    \centering
    \centering
    \includegraphics[width=\textwidth]{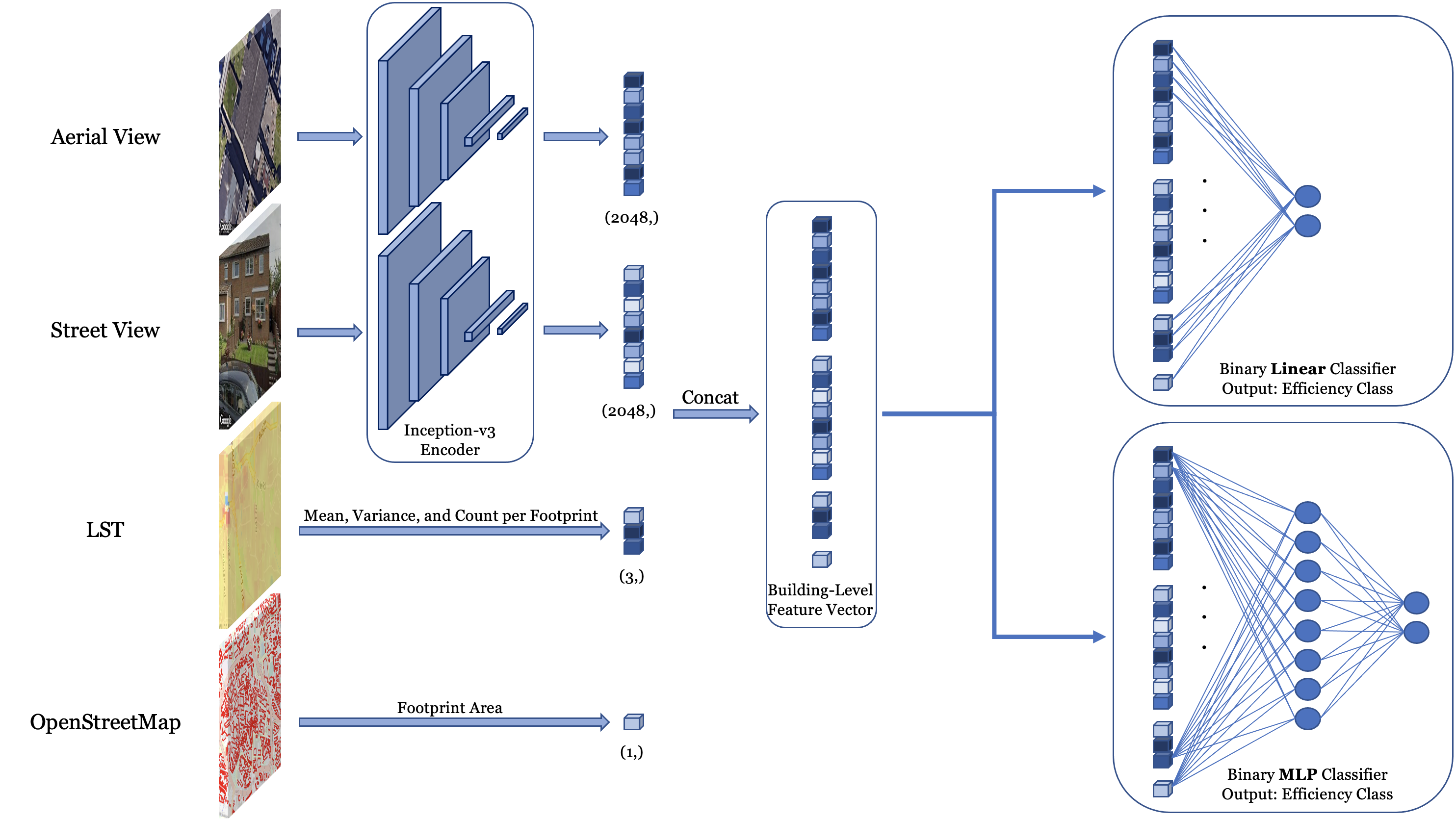}
    \caption{End-to-end deep learning approach.}
    \vspace{4pt}
    \label{fig:model_visualization}
\end{figure*}

Apart from the baseline models described in the previous section, this study proposes and evaluates an end-to-end deep learning architecture for predicting building energy efficiency from LST, footprint, street view and aerial imagery data as illustrated in Figure \ref{fig:model_visualization}. To do so, street view images and aerial images are each encoded with an Inception-v3-based encoder network \cite{inception}. These encoder networks have been pre-trained on the \textit{ImageNet} dataset \cite{imagenet} and their weights remain frozen. In contrast to the Inception-v3-based classification network, the feature encoder network drops the final affine layer and maps each image into a 2048-dimensional embedding space. In the embedding space, building-specific heat loss information and the building's footprint area are added as additional features and concatenated to the embedded street view and aerial image vectors. These building-level feature vectors are then either fed into a single- or multi-layer perceptron (MLP) in order to predict whether a given building is considered to be energy efficient (output is zero) or inefficient (output is one). While the single-layer prediction head contains significantly fewer parameters and is therefore less susceptible to overfitting, the MLP prediction head can use its ReLU non-linearity to learn a non-linear combination of the street view, aerial, LST, and footprint data. To reduce the risk of overfitting, the MLP prediction head contains only eight hidden neurons of which 50\% are randomly dropped during training \cite{dropout}. Unless specified otherwise, we generally refer to the model which fuses all four data sources in the MLP-based prediction head as the \textit{end-to-end deep learning} model.

\subsection{Evaluation metrics}

To evaluate the performance of our models in the binary classification setting, we use macro-averaged precision, recall, and F1 scores. In other words, we compute the individual class scores in terms of precision, recall, and F1 and average the results for each metric across classes. The rationale behind choosing macro-averaged scores is linked to the label imbalance in our dataset. As macro-averaged scores ensure that all classes receive equal weight during evaluation, macro-averaged scores are well-suited to evaluate algorithms on imbalanced datasets. In our case, this is important because the label distribution is imbalanced and can vary significantly between different cities. This is because each city is characterized by its unique history and the buildings vary significantly in terms of architectural style, age, and building materials, all of which influence the energy efficiency of the respective building stock. Thus, opting for macro-averaged evaluation metrics enables a more robust comparison between our models, even when faced with distribution shifts across different geographies.

\section{Results}

\subsection{Baselines}

For the k-NN baseline, the number of nearest neighbors is determined as $k = 3$ through a hyperparameter search on the validation set. Similarly, conducting a hyperparameter search on a logistic scale ranging from $C=1e^{-4}$ to $C=1000$, the SVM model with an inverse $\text{L}_2$-regularization strength of $C=1.0$ and a linear kernel is found to perform best on the validation data. For the SVM baseline, the same kernel values are re-used over multiple training iterations and saved in cache. As long as the resources needed for training stay within our hardware's memory requirements, the training time scales on the order of $O(n_{\text{features}} \cdot n_{\text{samples}}^2)$. However, once the cache is exhausted, the training time scales according to $O(n_{\text{features}} \cdot n_{\text{samples}}^3)$ which is why the SVM model is trained for a maximum of 10,000 iterations. Furthermore, the SVM model is trained with class weights that are inversely proportional to the class frequency in the training dataset.

\begin{table*}[h!]
\begin{center}
\caption{Quantitative model results on Peterborough test set.}
\label{table:baselines}
\begin{tabularx}{\textwidth}{lcccc}
\hline
& \multicolumn{4}{c}{Binary Classification (Macro)} \\
Model & Precision [\%] & Recall [\%] & F1 [\%] & $\Delta F1$ to MM [ppt] \\
\hline
Majority Model (MM) & 40.55 & 50.00 & 44.78 & - \\
k-Nearest Neighbor (k-NN) & 50.56 & 50.60 & 50.51 & +5.73 \\
Support Vector Machine (SVM) & 52.97 & 53.87 & 52.62 & +7.84 \\
\hline
End-to-End Deep Learning (MLP Head) & \textbf{68.30} & \textbf{63.05} & \textbf{64.64} & \textbf{+19.86} \\
\hline
\end{tabularx}
\end{center}
\end{table*}

As Table \ref{table:baselines} shows, the SVM classifier outperforms both, the k-NN model and the majority model across the macro-averaged precision, recall, and F1 scores. Notably, both the k-NN model and the SVM model perform significantly better than the majority model which speaks for the predictive potential of street view images as a data source. However, even the SVM-based model cannot exceed a marco-average F1 score of 52.62\%, leaving ample room for performance gains through potentially more sophisticated modeling techniques. We hypothesize that the SVM model performs best among all baselines in the binary classification setting because of its kernel-induced capability to effectively model data in high dimensions. The k-NN model, although being an extremely simple non-parametric algorithm, also performs significantly better than the majority model. This indicates that even pixel-by-pixel comparisons between street view images can be correlated with building energy efficiency. However, when validating the k-NN model and comparing its predictions on the Peterborough-based test set to the metrics obtained from the validation set in Coventry, Westminster, and Oxford, it becomes clear that the k-NN model is particularly susceptible to distribution shifts between cities, even within the United Kingdom. To put the performance of our baseline models into perspective, we compute the difference between each baseline and the majority model in terms of F1 score. While the k-NN and SVM-based models outperform the majority model by 5.73 and 7.84 percentage points respectively, the end-to-end deep learning model outperforms the majority model by 19.86 percentage points.

\subsection{End-to-end deep learning model}

As illustrated in Figure \ref{fig:model_visualization}, the end-to-end deep learning model fuses the different data sources in a multi-layer perceptron (MLP) predicition head and is thereby able to combine the signals from the different data sources non-linearly. The end-to-end deep learning model achieves a macro-averaged F1 score of 64.64\%, a precision of 68.30\%, and a recall of 63.05\%. In terms of F1 score, the end-to-end deep learning model outperforms the k-NN and SVM-based baseline models by 14.13 and 12.02 percentage points, respectively. The model hyperparameters are chosen via grid-search. The best performing model is trained with an Adam optimizer, a batch size of 16, a learning rate equal to $0.0001$, and class weights that are inversely proportional to the class frequency.

\subsection{Qualitative results}

\begin{figure*}[h!]
    \centering
    \begin{minipage}[b]{1.0\linewidth}
    \centering
    \begin{subfigure}[b]{0.45\textwidth}
        \centering
        \includegraphics[width=\textwidth]{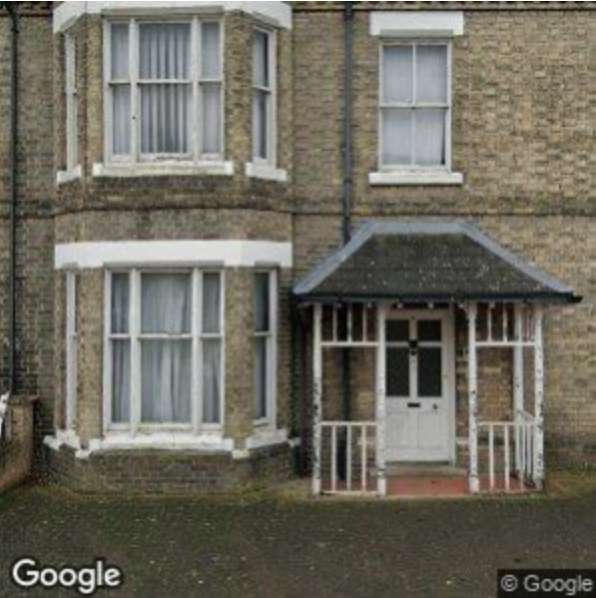}
        \caption{Original street view image}
        \label{fig:original_sv}
    \end{subfigure}
    \hspace{8pt}
    \begin{subfigure}[b]{0.45\textwidth}
        \centering
        \includegraphics[width=\textwidth]{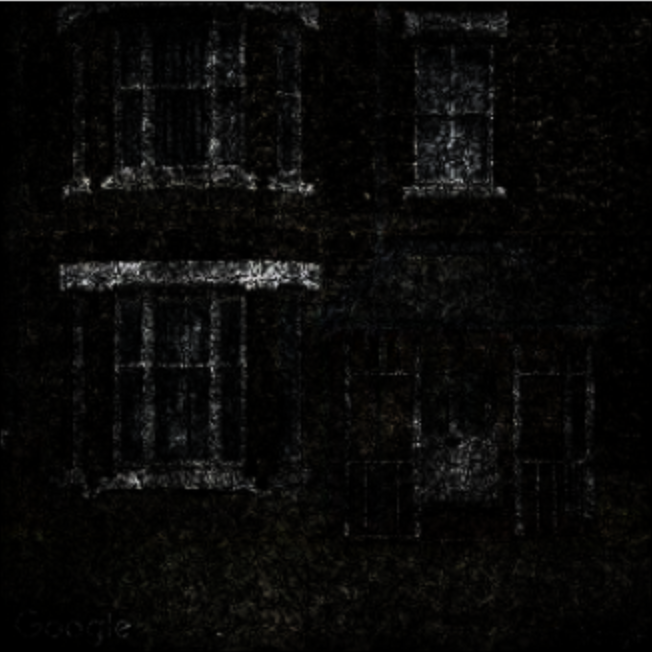}
        \caption{Attribution mask}
        \label{fig:street_img_attr}
    \end{subfigure}
    \caption{Street view attribution mask computed for a residential building in the test set.}
    \vspace{4pt}
    \label{fig:integrated-gradients-sv}
    \end{minipage}
\end{figure*}

Relying on the integrated gradients attribution method described in \cite{integrated-gradients}, we can increase the interpretability of our proposed end-to-end deep learning model in a qualitative manner. Building on \cite{kokhlikyan2020captum}, we use this method to quantify the joint and interdependent contributions of the street view and aerial images with respect to the final model's predictions. As described in \cite{integrated-gradients}, we use Equation \ref{eq:int-grad} to derive attribution maps for both, street view images and aerial images, in order to quantify the two image sources' impact on the difference in logit scores between the energy efficient and inefficient class.

\begin{equation}
    \text{Attr}(x_i) = (x_i - x^{\prime}) \cdot \int_{\alpha=0}^1 \frac{\partial F (x^{\prime} + \alpha \cdot (x_i - x^{\prime}))}{\partial x_i} d\alpha
    \label{eq:int-grad}
\end{equation}

$x_i$ represents the image for which we compute the attribution map, $F$ is the difference in the model's prediction of the efficient class logit score and the inefficient class logit score, and $x^{\prime}$ is a baseline image with respect to which the attribution map is generated. In our case, the baseline is simply an image with random pixels. We take the integral in Equation \ref{eq:int-grad} separately with respect to the street view image and the aerial image in order to generate an attribution map for each of the end-to-end model's image inputs. Since the definite integral in Equation \ref{eq:int-grad} is difficult to compute, it is approximated by a number of discrete intervals as shown in Equation \ref{eq:int-grad2} where we choose $m=50$.

\begin{equation}
    \text{Attr}(x_i) = (x_i - x^{\prime}) \cdot \frac{1}{m}\sum_{k=1}^m \frac{\partial F (x^{\prime} + \frac{k}{m} \cdot (x_i - x^{\prime}))}{\partial x_i}
    \label{eq:int-grad2}
\end{equation}

Figure \ref{fig:integrated-gradients-sv} juxtaposes an example of an original street view image and its attribution mask for a residential building located in the city of Peterborough, UK, which is part of our test set. We obtain the attribution mask from a pixel-wise multiplication of the original image with its normalized attribution values. The street view attribution mask in Figure \ref{fig:street_img_attr} reveals that the model puts significant weight on the building's windows and doors. This suggests that the model has learned to connect the position, shape, and size of windows and doors to a building's energy efficiency. In addition, the aerial attribution mask in Figure \ref{fig:aerial_img_attr} in the appendix suggests that the model starts to take roof geometries and a building's spatial context into consideration. As such, the visualized attribution masks might indicate that our end-to-end deep learning model has started to pick up on efficiency-related building features with respect to the size, shape, and position of windows. However, the visualized attribution masks also reveal that our models have not yet managed to develop a clear focus to identify efficiency-related features consistently. As a result, we cannot draw definite conclusions about whether our end-to-end deep learning model has learned to identify relevant parts of the street view and aerial images effectively. To make more definite claims, a larger dataset would be needed.

\subsection{Ablation study}

\begin{table*}[ht!]
\begin{center}
\caption{Test set results for ablation study.}
\label{table:ablation-study}
\begin{tabular}{lccc}
\hline
& \multicolumn{3}{c}{Binary Classification (Macro)} \\
Modeling Approach & Precision [\%] & Recall [\%] & F1 [\%] \\
\hline
Majority Model (Baseline) & 40.55 & 50.00 & 44.78 \\
\hline
\textbf{Energy Consumption + Log. Reg.} & \textbf{67.19} & \textbf{74.96} & \textbf{58.88} \\
Footprint Area + Log. Reg. & 57.95 & 55.91 & 56.37 \\
Street View Embedding + Log. Reg. & 61.13 & 55.21 & 55.35 \\
Aerial View Embedding + Log. Reg. & 59.73 & 53.75 & 53.15 \\
LST + Log. Reg. & 51.11 & 51.65 & 48.56 \\
\hline
End-to-End Deep Learning (MLP Head) & 68.30 & 63.05 & 64.64 \\
\hline
\end{tabular}
\end{center}
\end{table*}

In order to gain a better understanding of the predictive power of each data source, we conduct an ablation study in which we compare the performance of the majority model to logistic regression-based models trained only on a single data source. The results of the ablation study are shown in Table \ref{table:ablation-study}. While the first column specifies the respective modeling approach, the subsequent columns report each model's performance in terms of macro-averaged precision, recall, and F1 scores. The rows are sorted in descending order with respect to the F1 scores and the best performing logistic regression model is printed in bold. It is important to note that apart from the remotely sensed data sources, namely street view, aerial view, LST, and footprint data, this comparison also includes a logistic regression model trained on energy consumption data (kWh/$m^2$). This is because industry experts generally use energy consumption data as a proxy for the energy efficiency of a building. As a result, this ablation study can help us understand how linear classifiers trained on each remotely sensed data source perform in comparison to a majority model and a linear classifier trained on energy consumption data. To ensure the comparability of the results, every logistic regression-based model uses the scikit-learn implementation \cite{scikit-learn} and is trained with stochastic gradient descent as an optimizer, a common seed value, and no class weights.

The ablation study reveals that all logistic regression-based models outperform the majority model. This means that every data source contains a predictive signal which can be used to classify buildings in terms of energy efficiency. As expected, the model trained on energy consumption data performs best and achieves an F1 score of 58.88\%. In other words, the model trained on energy consumption data outperforms the next best competitor models by 6.06 percentage points in terms of macro-averaged precision, by 19.05 percentage points in terms of macro-averaged recall, and by 2.51 percentage points in terms of macro-averaged F1. Among the remotely sensed data sources, the logistic regression-based model trained on footprint area performs best, achieving a macro-averaged F1 score of 56.37\%. Comparing the footprint area model to the image-based models, we can see that the model trained on street view embeddings achieves similar results, whereas the model trained on aerial view embeddings exhibits slightly lower macro-averaged recall and F1 scores at 53.75\% and 53.15\%, respectively. Lastly, the model trained on LST data achieves an F1 score of only 48.56\%, the lowest among all single-data-source models and only 3.78 percentage points higher than the majority model. Intuitively, this makes sense as the the signal contained in the LST data is, until now, limited due to its coarse spatial resolution of 30x30m. 
Hence, we can conclude that while all remotely sensed data sources exhibit signals to predict building energy efficiency, energy consumption data appears to be the most predictive feature.

Having established that all presented data sources contain signals in terms of building energy efficiency, we continue by examining the predictive potential of combining these data sources. To do so, we conduct a second ablation study in which we train our proposed end-to-end deep learning architecture, as described in Figure \ref{fig:model_visualization}, on different feature combinations. Unlike the logistic regression-based models in Table \ref{table:ablation-study} which are restricted to draw a linear decision boundary, this second ablation study also highlights the differences in model performance which arise from combining the presented data sources linearly in a single-layer perceptron as well as non-linearly in a multi-layer perceptron prediction head. The results of this end-to-end deep learning ablation study are presented in Table \ref{tab:e2e-ablation-study}. To ensure the comparability of the results, all models are trained in PyTorch \cite{pytorch} with an Adam optimizer \cite{adam}, a batch size of 16, a learning rate equal to $0.0001$, and class weights that are inversely proportional to the class frequency in the training dataset. 

\begin{table}[h!]

    \caption{Test set results for end-to-end deep learning ablation study.}
    \label{tab:e2e-ablation-study}
    \begin{subtable}[h]{\textwidth}
    \centering
    \caption{Linear prediction head.}
    \label{tab:linear-ablation}
        \footnotesize

    \begin{tabular}{lrrr}
    \hline
    & \multicolumn{3}{c}{Binary Classification (Macro)} \\
    Input Features & Precision [\%] & Recall [\%] & F1 Score [\%] \\
    \hline
    \textbf{Aerial View (AV)} & \textbf{64.42} & \textbf{58.59} & \textbf{59.68} \\
    Street View (SV) & 60.20 & 57.79 & 58.47 \\
    Segmented SV & 61.54 & 56.40 & 56.98 \\
    Energy Consumption (EC) & 66.51 & 73.42 & 56.72 \\
    Footprint (FP) & 52.86 & 54.35 & 48.45 \\
    Land Surface Temperature (LST) & 60.49 & 50.65 & 18.62 \\
    \hline
    \textbf{AV, FP} & \textbf{64.35} & \textbf{62.96} & \textbf{63.55} \\
    AV, LST & 65.36 & 61.36 & 62.58 \\
    SV, AV & 66.83 & 60.32 & 61.77 \\
    SV, FP & 59.61 & 62.22 & 60.09 \\
    SV, LST & 58.45 & 59.67 & 58.86 \\
    LST, FP & 60.48 & 50.60 & 18.51 \\
    \hline
    \textbf{SV, AV, LST} & \textbf{64.30} & \textbf{63.92} & \textbf{64.10} \\
    SV, AV, FP & 66.01 & 62.48  & 63.68 \\
    AV, LST, FP & 60.89 & 63.20 & 61.54 \\
    SV, LST, FP & 60.33 & 60.33 & 60.42 \\
    \hline
    \textbf{SV, AV, LST, FP} & \textbf{62.81} & \textbf{63.61}  & \textbf{63.17} \\
    \hline
    \textbf{SV, AV, LST, FP, EC} & \textbf{78.89} & \textbf{86.65} & \textbf{81.44} \\
    \hline
    \multicolumn{4}{r}{observations = 3,700}\\
    \end{tabular}
    \end{subtable}
    
    \begin{subtable}[h]{\textwidth}
    \centering
    \caption{MLP prediction head.}
    \label{tab:mlp-ablation}
        \footnotesize

    \begin{tabular}{lrrr}
    \hline
    & \multicolumn{3}{c}{Binary Classification (Macro)} \\
    Input Features & Precision [\%] & Recall [\%] & F1 Score [\%] \\
    \hline
    \textbf{Aerial View (AV)} & \textbf{64.31} & \textbf{63.69} & \textbf{63.98} \\
    Street View (SV) & 60.64 & 60.67 & 60.66 \\
    Segmented SV & 59.91 & 57.95 & 58.57 \\
    Energy Consumption (EC) & 65.21 & 70.04 & 51.95 \\
    Footprint (FP) & 53.21 & 54.88 & 49.52 \\
    Land Surface Temperature (LST) & 60.49 & 50.67 & 18.66 \\
    \hline
    \textbf{SV, AV} & \textbf{67.93} & \textbf{63.29} & \textbf{64.78} \\
    AV, FP & 63.97 & 63.59 & 63.77 \\
    AV, LST & 62.17 & 62.90 & 62.50 \\
    SV, FP & 62.94 & 59.94 & 60.88 \\
    SV, LST & 61.55 & 59.41 & 60.13 \\
    LST, FP & 60.48 & 50.58 & 18.47 \\
    \hline
    \textbf{AV, LST, FP} & \textbf{63.13} & \textbf{64.11} & \textbf{63.56} \\
    SV, AV, LST & 69.63 & 61.35  & 63.14 \\
    SV, AV, FP & 69.61 & 60.86 & 62.59 \\
    SV, LST, FP & 63.02 & 61.11 & 61.84 \\
    \hline
    \textbf{SV, AV, LST, FP} & \textbf{68.30} & \textbf{63.05} & \textbf{64.64} \\
    \hline
    \textbf{SV, AV, LST, FP, EC} & \textbf{74.91} & \textbf{86.33} & \textbf{76.09} \\
    \hline
    \multicolumn{4}{r}{observations = 3,700}\\
    \end{tabular}
    \end{subtable}

\end{table}

While the first column in Table \ref{tab:e2e-ablation-study} specifies the respective combination of input features for our end-to-end deep learning model, the subsequent columns report the model performance in terms of macro-averaged precision, recall, and F1 scores. Table \ref{tab:linear-ablation} reports the results for the linear prediction head and Table \ref{tab:mlp-ablation} for the MLP-based prediciton head. To improve the readability, we sort the rows in each sub-group with respect to their F1 scores and print the row with the highest F1 score per sub-group in bold. Apart from the models trained on different combinations of remotely sensed data sources, we also report the performance of two additional models for each prediction head, one trained on energy consumption data only and another trained on all remotely sensed data sources plus energy consumption. Juxtaposing the experiments in Table \ref{table:ablation-study} and Table \ref{tab:linear-ablation}, it is important to note that the main difference between both modeling approaches is the introduction of class weights. This means that if we were to train the linear prediction head in Table \ref{tab:linear-ablation} without a class weighted loss, the results would be almost identical to those presented in Table \ref{table:ablation-study}.

When comparing the linear with the MLP-based prediction head, we can see that the performance differences between the two approaches are only marginal. For both approaches, the best performing models in each sub-group achieve macro-averaged F1 scores between 63.17\% and 64.78\%. Only in the single-data-source case does the performance of the linear prediction head fall out of this range, scoring 59.68\% in terms of F1 with aerial imagery as its input feature. This difference in performance is likely caused by the fact that the image-based data sources exhibit a more non-linear signal than the LST, footprint area, and energy consumption features and therefore benefit more from a model which can combine information in a non-linear fashion. Similarly, both approaches achieve comparable results in terms of macro-averaged precision and recall. Hence, we can conclude that adding a hidden layer with a ReLU non-linearity to the linear prediction head does only improve prediction performance slightly, with the biggest performance improvements observed for purely image-based input data sources.

In Section \ref{data_cleaning_segmentation}, we described an optional data processing step in which we segmented out the sky from street view imagery. With the ablation study, we can examine whether this processing step improves the model's prediction performance. To do so, we train one end-to-end deep learning model on the original street view imagery and one on the segmented street view imagery for each prediction head. In both cases, we do not see an increase in prediction performance with the segmented street view images. This might be caused by the fact that the ImageNet-based Inception-v3 encoder has never seen segmented images in its training dataset. Hence, by segmenting out the sky, we introduce an artificial and unwanted distribution shift which hinders the pre-trained Inception-v3 model from encoding its input images effectively. As a result, we recommend that this processing step should not be considered in subsequent studies.

When training the linear and MLP-based prediction heads on LST data only, we observe very low macro-averaged F1 scores in the range of 18.62-18.66\%. With F1 scores in this range, we can conclude that both models did not succeed in learning from the signal in the LST data. Unlike the logistic regression-based approach in Table \ref{table:ablation-study} which achieves a macro-averaged F1 score of 48.56\% with LST as its single input feature, the linear and MLP-based prediction heads converge to sub-optimal local minima during training. From our experiments, we know that these undesirable local minima are caused by the introduction of class weights during training. However, to improve the comparability of the models in this ablation study, we choose to train every model with the same class weights, even if this means that some models will fail to learn effectively.

Among the end-to-end deep learning models trained on remotely sensed data sources, we can see that only two models achieve macro-averaged F1 scores larger than 64.50\% on the test set, both of them relying on MLP-based prediction heads. While both models achieve very similar scores across all macro-averaged metrics, it is interesting to note that the best performing model in terms of F1 score relies exclusively on image-based data sources, in this case street view and aerial view imagery. Considering the results in Table \ref{table:ablation-study}, this result might be counter-intuitive at first as we found that the linear classifier trained on footprint area data performed best among the classifiers relying on only one remotely sensed data source. On second thought, this apparent mismatch resolves as we realize that the image-based data sources do not only contain information about a building's footprint area, as shown in \cite{Streltsov2020}, but also many other features which we expect to correlate with a building's energy efficiency. As an example, street view as well as aerial images provide information on the spatial context of a building, such as the amount of vegetation, the make, age, and size of cars, and the building density, all of which are indicative of the socio-economic status for a given neighborhood and thus correlate with energy efficiency. Similarly, street view and aerial images provide information on the size, age, and condition of a building's windows, walls, and roof, i.e. features which are directly linked to and drive the building's rating in terms of energy efficiency. Hence, it is not surprising to see in Table \ref{tab:linear-ablation} that the linear models trained on a single image-based data source do indeed outperform all other single-data-source linear classifiers, even those in Table \ref{table:ablation-study}. As a result, we can conclude that estimating building energy efficiency from remotely sensed data sources greatly benefits from image-based input features. Since the performance gap between the classifiers trained on a single image-based data source and those trained on a single non-image-based data source only widens when we examine the results of the MLP-based prediction heads in Table \ref{tab:mlp-ablation}, we can also conclude that the image-based features exhibit a non-linear decision boundary with respect to building energy efficiency and should therefore be fused in an MLP-based prediction head. While footprint area and LST data carry less predictive signal than the image-based data sources, both still provide meaningful insights to improve prediction performance and together can replace one of the image-based data sources without a significant loss in performance. As such, the ablation study shows the complementary nature of our data sources and the benefits from fusing them in a single model. For these reasons, we suspect that by increasing the dataset size, the MLP-based prediction head trained on all remotely sensed data sources should be able to outperform the MLP head trained exclusively on images.

In addition, the ablation study underscores the role and value of energy consumption data as a predictor for building energy efficiency. While the logistic regression-based model trained on energy consumption data achieves a precision score of 67.19\%, a recall score of 74.96\%, and an F1 score of 58.88\% (Figure \ref{table:ablation-study}), our end-to-end deep learning models which are trained on a combination of energy consumption and remotely sensed data sources perform significantly better. Fusing street view, aerial view, footprint, LST, and energy consumption data, our linear prediction head is able to achieve a precision score of 78.89\%, a recall score of 86.65\%, and an F1 score of 81.44\%. This means that by incorporating energy consumption data as a feature, the linear prediction head is able to improve the performance of the MLP-based prediction head trained on all remotely sensed input features by 15.51\% in terms of precision, by 37.45\% in terms of recall, and by 26\% in terms of F1 score. Hence, we recommend to add household-level energy consumption data as a feature whenever the data is available. However, even in the absence of household-level energy consumption data, our end-to-end deep learning model trained exclusively on remotely sensed data sources is able to provide critical insights in a scalable and non-intrusive way.

\section{Discussion and conclusion}

We presented a novel way to estimate building energy efficiency, relying only on remotely sensed data sources. To do so, we developed an end-to-end deep learning model and explored the predictive power of multiple widely available data sources in isolation as well as in combination. As a result, our contribution is three-fold.

First, we created a novel dataset representing almost 40,000 buildings across four diverse regions in the United Kingdom in terms of street view, aerial view, footprint, and LST data. Confronted with the challenge of reducing noise in the individual data sources, we decided to average the LST raster data across multiple dates, automatically remove empty, i.e. "Sorry, we have no imagery here", aerial images according to a characteristic pixel combination, and clean street view images by identifying noisy clusters in an embedding space. As a result, we were able to increase the dataset's quality and its geographic scope across multiple geographies while minimizing manual data labeling efforts.  

Second, we contributed an assessment of ANN-based methods to classify buildings as energy efficient (EU rating A-D) or inefficient (EU rating E-G).
While we developed the class definitions together with industry experts, the final dataset exhibits a strong class imbalance which hampers model training and the evaluation of results. Hence, we decided to evaluate the model performance qualitatively with the integrated gradients method and quantitatively in terms of macro-averaged precision, recall, and F1 scores. By combining the qualitative and quantitative evaluation approaches, we were able to identify and understand shortcomings in the current models more intuitively. For example, the integrated gradients method revealed that the deep learning models have not yet learned to consistently pay attention to efficiency-related image features.

We concluded this work by performing an ablation study in which we examined the potential of different feature subsets and model architectures in terms of prediction performance. Our results indicate that in the binary setting of predicting building energy efficiency, the end-to-end deep learning model achieves a macro-averaged F1 score of 64.64\%. When comparing the performance of the end-to-end deep learning model with the baseline models and the models trained in the ablation study, we can conclude that the end-to-end deep learning model has a superior performance to all but one approach. The approach which achieves the best performance in terms of F1 score involves fusing street view and aerial view imagery with the MLP-based prediction head of our end-to-end deep learning architecture. As this model ablation only adds 0.14 percentage points in comparison to the model trained on all remotely sensed data sources, we are confident that a larger dataset size would enable the latter model to learn an even more powerful feature combination and consistently outperform all model ablations.

In summary, this paper presents a novel and non-intrusive approach to predict building energy efficiency from remotely sensed data sources only. Using a new combination of input data sources to train multiple end-to-end deep learning models, we are able to achieve a macro-averaged precision of 68.30\%, a recall of 63.05\%, and an F1 score of 64.64\% in our binary classification setting. While these results indicate that our current approach is limited in its ability to predict building energy efficiency, the proposed approach enables a fast and scalable analysis of the existing building stock. Furthermore, based on the results in \cite{Streltsov2020}, we are confident that averaging the model's predictions across neighborhoods improves the model performance significantly. Hence, the current model can already provide meaningful insights to decision makers when prioritizing districts by their retrofit potential. 

In terms of limitations, we argue that the model's limited prediction performance is mainly driven by three factors. First, to train an effective classifier, large yet clean datasets are preferred. As a result, we expect that our model's performance could be improved by increasing the current dataset size of 39,605 observations. Second, the spatial resolution of the LST data hampers prediction performance. While our ablation study has shown that LST measurements are indeed a predictive feature to estimate building energy efficiency (see Table \ref{table:ablation-study}), the models trained exclusively on LST data perform worst among all single-data-source models. This is because, at its current spatial resolution of 30 by 30 meters, the LST data provides only a relatively weak and noisy signal with respect to building energy efficiency. However, with the increasing availability of higher resolution thermal infrared imagery, LST measurements are likely to become a key feature in predicting building energy efficiency soon. Furthermore, the predictive power of the four data sources used in this study is inherently limited when it comes to predicting building energy efficiency. This is because building energy efficiency is a highly complex metric and therefore difficult to estimate by analyzing buildings from the outside only. Nevertheless, our ablation study shows the complementary nature of the presented data sources. This indicates that adding other remotely sensed data sources might further improve prediction performance. The ablation study also shows that including a household's energy consumption as a feature boosts prediction performance significantly, improving the performance of the original end-to-end deep learning model by 15.51\% in terms of precision, by 37.45\% in terms of recall, and by 26\% in terms of F1 score. As a result, adding household-level energy consumption data as a feature is recommended whenever the data is available.

Future work could continue to experiment with the proposed end-to-end deep learning architecture in order to increase prediction performance. Furthermore, future work could examine the potential of LST-based data to estimate household-level energy consumption. Lastly, based on our results which highlight the complementary nature of different remotely sensed data sources, integrating other data modalities seems promising. As an example, Lidar-derived point clouds could be a sensible addition in order to improve the model's performance while ensuring the scalability of the proposed method across different countries in the EU and beyond. In combination with the increasing availability and resolution of existing data sources, we think that this method can play an important role in transforming the building sector by providing critical information to retrofit markets fast, accurately, and across geographies in a non-intrusive manner.

\section{Credits and acknowledgements}

\noindent \textit{Kevin Mayer}: Conceptualization, Methodology, Software, Validation, Formal analysis, Investigation, Visualization, Data Curation, Writing – Original Draft, Writing – Review and Editing.\\
\textit{Lukas Haas}: Methodology, Software, Formal analysis, Visualization, Data Curation, Writing – Original Draft.\\
\textit{Tianyuan Huang}: Methodology, Data Curation, Software, Visualization.\\
\textit{Juan Bernabé-Moreno}: Supervision, Resources.\\
\textit{Ram Rajagopal}: Supervision, Project Administration, Funding.\\
\textit{Martin Fischer}: Supervision, Project Administration, Funding, Resources, Writing – Review and Editing.\\
\\
\noindent This research was supported by Stanford's Bits\&Watts initiative in collaboration with \textit{E.ON SE}. Kevin Mayer would like to express his gratitude to Stefan Padberg who has been a great advocator and ambassador for this research. Furthermore, Kevin Mayer would like to thank Dr. Alejandro Newell for many insightful discussions. Kevin Mayer would also like to thank Isabel Larus and Chris Agia for proofreading and giving feedback on the manuscript.

\newpage
\section{Appendix}

\subsection{K-Means cluster analysis}

\begin{figure*}[h!]
    \centering
    \begin{minipage}[b]{1.0\linewidth}
    \centering
    \begin{subfigure}[b]{0.45\textwidth}
        \centering
        \includegraphics[width=\textwidth]{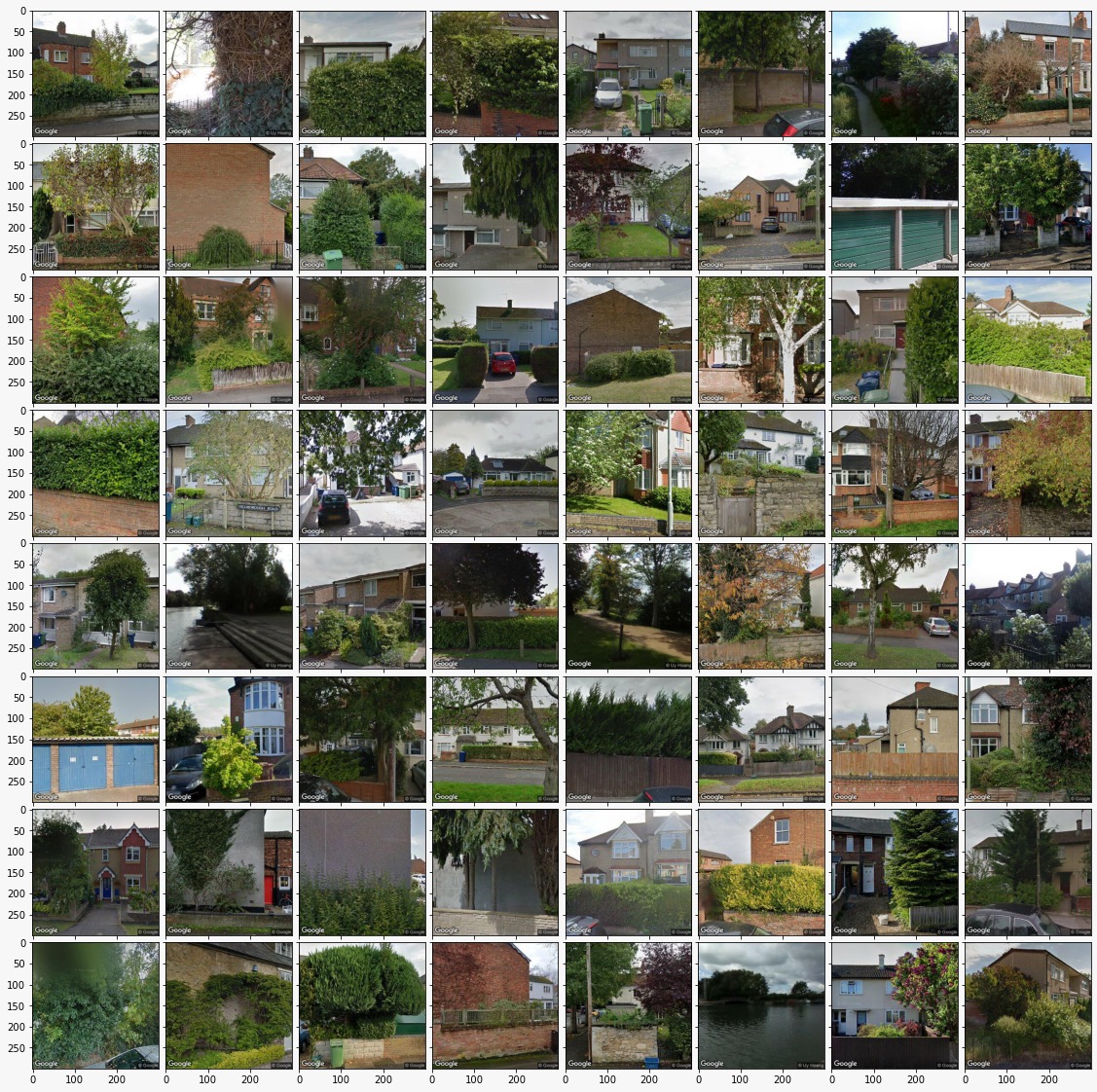}
        \caption{Suburban street views}
    \end{subfigure}
    \hspace{8pt}
    \begin{subfigure}[b]{0.45\textwidth}
        \centering
        \includegraphics[width=\textwidth]{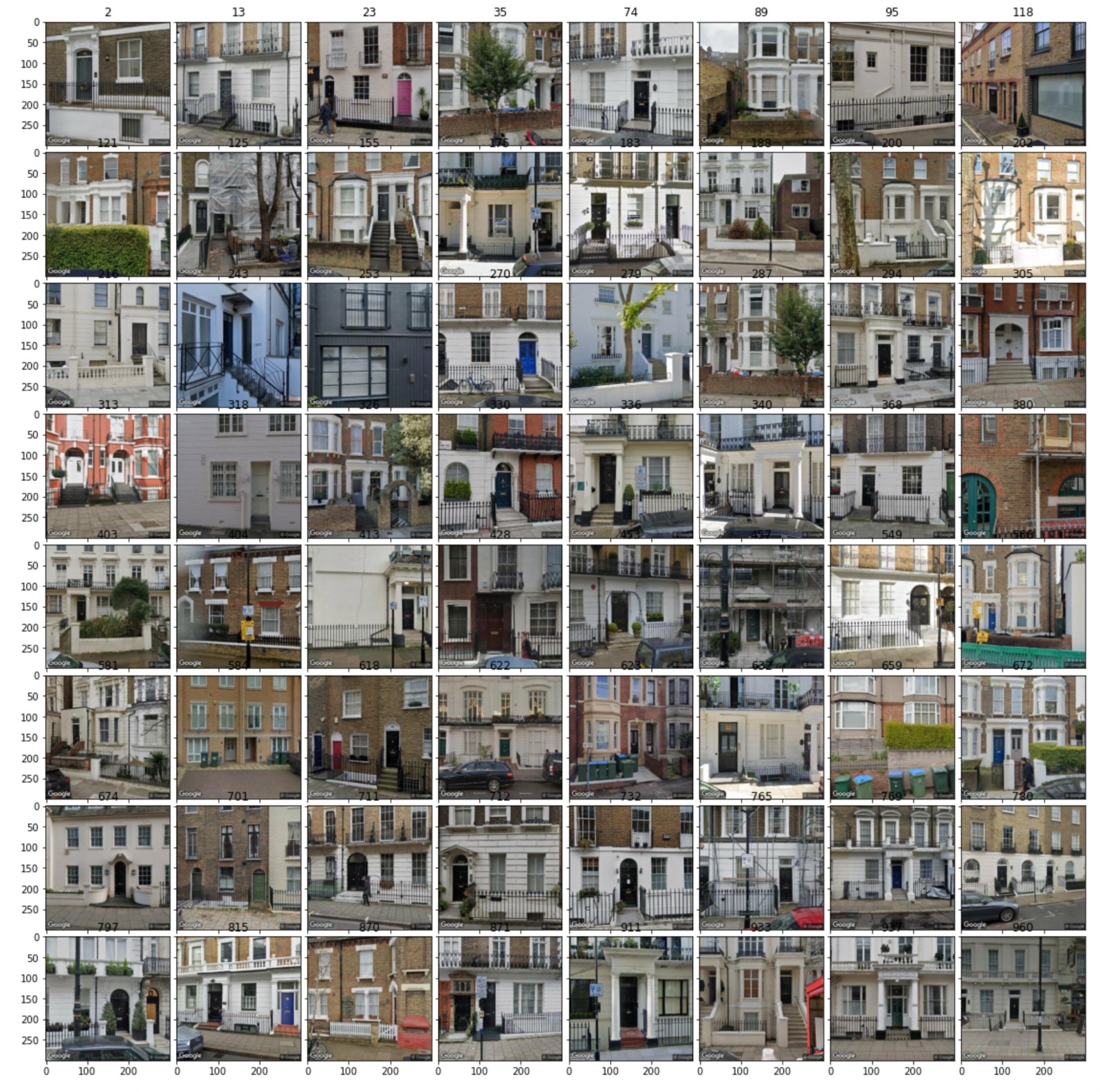}
        \caption{Urban street views}
    \end{subfigure}
    \caption{Clusters representing suburban and urban street views.}
    \vspace{4pt}
    \label{fig:architectural_styles}
    \end{minipage}
\end{figure*}

Figure \ref{fig:architectural_styles} juxtaposes two of the clusters which we identified by clustering the street view embeddings with K-Means as described in Section \ref{data_cleaning_segmentation}. In addition to Figure \ref{fig:clusters}, this example highlights how well the K-Means clustering algorithm is able to increase the similarity of samples in a given cluster while maximizing the sample differences between clusters. While the cluster on the left predominantly depicts residential brick buildings in a suburban environment rich in vegetation, the cluster on the right depicts a more urbanized environment with buildings in a Victorian style. As a result, clustering street view image embeddings enables us to group buildings in terms of their architectural styles as well as their immediate built environment.

\subsection{Qualitative results: aerial view attribution mask}

\begin{figure*}[h!]
    \centering
    \begin{minipage}[b]{1.0\linewidth}
    \centering
    \begin{subfigure}[b]{0.45\textwidth}
        \centering
        \includegraphics[width=\textwidth]{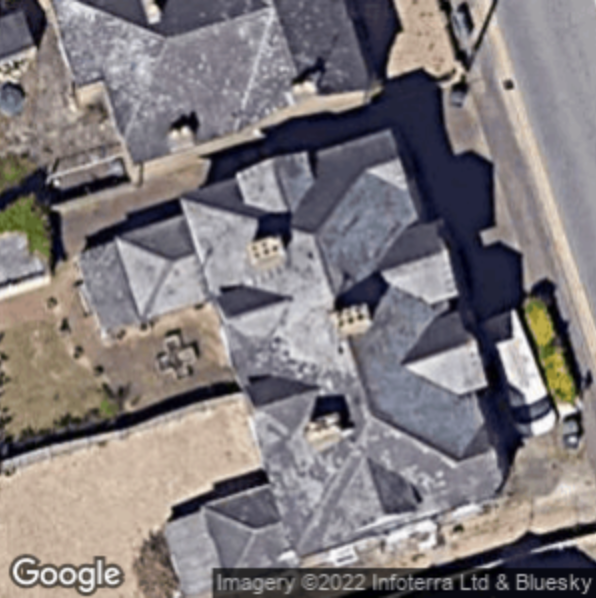}
        \caption{Original aerial view image}
        \label{fig:original_aerial_img}
    \end{subfigure}
    \hspace{8pt}
    \begin{subfigure}[b]{0.45\textwidth}
        \centering
        \includegraphics[width=\textwidth]{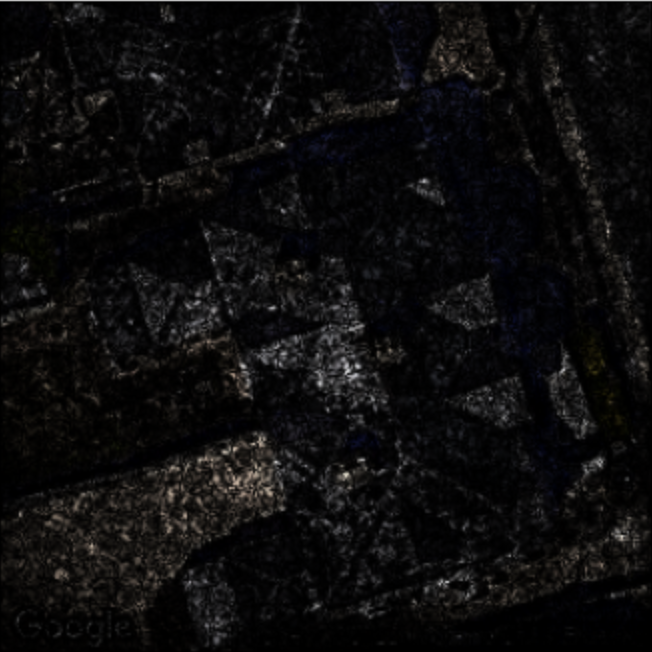}
        \caption{Attribution mask}
        \label{fig:aerial_img_attr}
    \end{subfigure}
    \caption{Aerial view attribution mask computed for a residential building in the test set.}
    \vspace{4pt}
    \label{fig:integrated-gradients-av}
    \end{minipage}
\end{figure*}

Inspecting the aerial attribution image in Figure \ref{fig:aerial_img_attr}, we can see that the model focuses mostly on two parts of the original aerial image, i.e. the roof of the building in the center and the backyard at the bottom left. In addition, the model puts emphasis on the pixels along the building front yards. Hence, the aerial attribution mask might suggest that the model starts to consider a building's outline, its spatial context, as well as its roof geometry when classifying energy efficiency. While we are able to observe a number of promising patterns in the attribution mask, it is important to note that the attribution mask does still exhibit a significant amount of noise. This suggests that our model has not yet learned to identify building energy efficiency related characteristics consistently.

\newpage
\singlespacing
\bibliographystyle{ieeetr} 
\bibliography{literature}

\begin{thebibliography}{10}

\bibitem{EuDirective}
{European Commission}, ``Energy efficiency in buildings.''
  \url{https://ec.europa.eu/info/news/focus-energy-efficiency-buildings-2020-lut-17_en},
  2020.
\newblock Accessed: 2022-04-29.

\bibitem{IEA2019}
{International Energy Agency}, ``The critical role of buildings.''
  \url{https://www.iea.org/reports/the-critical-role-of-buildings}, 2019.
\newblock Accessed: 2022-07-26.

\bibitem{dupont}
N.~Milojevic-Dupont and F.~Creutzig, ``Machine learning for geographically
  differentiated climate change mitigation in urban areas,'' {\em Sustainable
  Cities and Society}, vol.~64, p.~102526, 2021.

\bibitem{EPC}
{Housing \& Communities Department for Levelling Up}, ``Energy performance of
  buildings data: England and wales.''
  \url{https://epc.opendatacommunities.org/}, 2021.
\newblock Accessed: 2022-04-15.

\bibitem{Pham2020}
A.~D. Pham, N.~T. Ngo, T.~T.~H. Truong, N.~T. Huynh, and N.~S. Truong,
  ``Predicting energy consumption in multiple buildings using machine learning
  for improving energy efficiency and sustainability,'' {\em Journal of Cleaner
  Production}, vol.~260, 7 2020.

\bibitem{Streltsov2020}
A.~Streltsov, J.~M. Malof, B.~Huang, and K.~Bradbury, ``Estimating residential
  building energy consumption using overhead imagery,'' {\em Applied Energy},
  vol.~280, 12 2020.

\bibitem{Dougherty2021}
T.~R. Dougherty, T.~Huang, Y.~Chen, R.~K. Jain, and R.~Rajagopal, ``Schmear:
  Scalable construction of holistic models for energy analysis from rooftops,''
  pp.~111--120, Association for Computing Machinery, Inc, 11 2021.

\bibitem{Rosenfelder2021}
M.~Rosenfelder, M.~Wussow, G.~Gust, R.~Cremades, and D.~Neumann, ``Predicting
  residential electricity consumption using aerial and street view images,''
  {\em Applied Energy}, vol.~301, 11 2021.

\bibitem{Kontokosta2012}
C.~E. Kontokosta, ``Predicting building energy efficiency using new york city
  benchmarking data,'' 2012.

\bibitem{Sun2022}
M.~Sun, C.~Han, Q.~Nie, J.~Xu, F.~Zhang, and Q.~Zhao, ``Understanding building
  energy efficiency with administrative and emerging urban big data by deep
  learning in glasgow,'' {\em Energy and Buildings}, p.~112331, 2022.

\bibitem{Lee2019}
S.~Lee, S.~Iyengar, M.~Feng, P.~Shenoy, and S.~Maji, ``Deeproof: A data-driven
  approach for solar potential estimation using rooop imagery,''
  pp.~2105--2113, Association for Computing Machinery, 7 2019.

\bibitem{Krapf2021}
S.~Krapf, N.~Kemmerzell, S.~K.~H. Uddin, M.~H. Vázquez, F.~Netzler, and
  M.~Lienkamp, ``Towards scalable economic photovoltaic potential analysis
  using aerial images and deep learning,'' {\em Energies}, vol.~14, 7 2021.

\bibitem{DSfG}
K.~Mayer, Z.~Wang, M.-L. Arlt, D.~Neumann, and R.~Rajagopal, ``Deepsolar for
  germany: A deep learning framework for pv system mapping from aerial
  imagery,'' in {\em 2020 International Conference on Smart Energy Systems and
  Technologies (SEST)}, pp.~1--6, 2020.

\bibitem{Rausch2020}
B.~Rausch, K.~Mayer, M.-L. Arlt, G.~Gust, P.~Staudt, C.~Weinhardt, D.~Neumann,
  and R.~Rajagopal, ``{An Enriched Automated PV Registry: Combining Image
  Recognition and 3D Building Data},'' in {\em 34th Conference on Neural
  Information Processing Systems (NeurIPS 2020)}, 2020.

\bibitem{Mayer2022}
K.~Mayer, B.~Rausch, M.~L. Arlt, G.~Gust, Z.~Wang, D.~Neumann, and
  R.~Rajagopal, ``3d-pv-locator: Large-scale detection of rooftop-mounted
  photovoltaic systems in 3d,'' {\em Applied Energy}, vol.~310, 3 2022.

\bibitem{Hoffmann2019}
E.~J. Hoffmann, Y.~Wang, M.~Werner, J.~Kang, and X.~X. Zhu, ``Model fusion for
  building type classification from aerial and street view images,'' {\em
  Remote Sensing}, vol.~11, 6 2019.

\bibitem{Bin2020}
J.~Bin, B.~Gardiner, E.~Li, and Z.~Liu, ``Multi-source urban data fusion for
  property value assessment: A case study in philadelphia,'' {\em
  Neurocomputing}, vol.~404, pp.~70--83, 9 2020.

\bibitem{footprintextraction}
W.~Li, C.~He, J.~Fang, J.~Zheng, H.~Fu, and L.~Yu, ``Semantic
  segmentation-based building footprint extraction using very high-resolution
  satellite images and multi-source gis data,'' {\em Remote Sensing}, vol.~11,
  no.~4, p.~403, 2019.

\bibitem{Deb2021}
C.~Deb, Z.~Dai, and A.~Schlueter, ``A machine learning-based framework for
  cost-optimal building retrofit,'' {\em Applied Energy}, vol.~294, 7 2021.

\bibitem{berrill2022}
P.~Berrill, E.~Wilson, J.~Reyna, A.~Fontanini, and E.~Hertwich,
  ``Decarbonization pathways for the residential sector in the united states,''
  {\em Nature Climate Change}, 8 2022.

\bibitem{Tsanas2012}
A.~Tsanas and A.~Xifara, ``Accurate quantitative estimation of energy
  performance of residential buildings using statistical machine learning
  tools,'' {\em Energy and Buildings}, vol.~49, pp.~560--567, 6 2012.

\bibitem{Chou2018}
J.~S. Chou and D.~S. Tran, ``Forecasting energy consumption time series using
  machine learning techniques based on usage patterns of residential
  householders,'' {\em Energy}, vol.~165, pp.~709--726, 12 2018.

\bibitem{Gorelick2017}
N.~Gorelick, M.~Hancher, M.~Dixon, S.~Ilyushchenko, D.~Thau, and R.~Moore,
  ``Google earth engine: Planetary-scale geospatial analysis for everyone,''
  {\em Remote Sensing of Environment}, vol.~202, pp.~18--27, 12 2017.

\bibitem{Ermida2020}
S.~L. Ermida, P.~Soares, V.~Mantas, F.~M. Göttsche, and I.~F. Trigo, ``Google
  earth engine open-source code for land surface temperature estimation from
  the landsat series,'' {\em Remote Sensing}, vol.~12, 5 2020.

\bibitem{OSM}
{OpenStreetMap contributors}, ``Openstreetmap.''
  \url{https://www.openstreetmap.org}, 2017.
\newblock Accessed: 2022-04-15.

\bibitem{GoogleCloud}
{Google Cloud}, ``Google cloud platform.'' \url{https://cloud.google.com/}.
\newblock Accessed: 2022-04-15.

\bibitem{inception}
C.~Szegedy, V.~Vanhoucke, S.~Ioffe, and J.~Shlens, ``Rethinking the inception
  architecture for computer vision,''

\bibitem{socioeconomicsGSV}
J.~Machicao, A.~Specht, D.~Vellenich, L.~Meneguzzi, R.~David, S.~Stall,
  K.~Ferraz, L.~Mabile, M.~O'brien, and P.~Corr{\^e}a, ``{A Deep-Learning
  Method for the Prediction of Socio-Economic Indicators from Street-View
  Imagery Using a Case Study from Brazil},'' {\em {CODATA Data Science
  Journal}}, vol.~21, Feb. 2022.

\bibitem{buildingageGSV}
Y.~Li, Y.~Chen, A.~Rajabifard, K.~Khoshelham, and M.~Aleksandrov, ``{Estimating
  Building Age from Google Street View Images Using Deep Learning (Short
  Paper)},'' in {\em 10th International Conference on Geographic Information
  Science (GIScience 2018)}, vol.~114, pp.~40:1--40:7, Schloss
  Dagstuhl--Leibniz-Zentrum fuer Informatik, 2018.

\bibitem{inplace_abn}
Mapillary, ``inplace-abn github repository.''
  \url{https://github.com/mapillary/inplace_abn}, 2022.
\newblock Accessed: 2022-03-14.

\bibitem{Cityscapes}
M.~Cordts, M.~Omran, S.~Ramos, T.~Rehfeld, M.~Enzweiler, R.~Benenson,
  U.~Franke, S.~Roth, B.~Schiele, D.~A.R\&D, and T.~U. Darmstadt, ``The
  cityscapes dataset for semantic urban scene understanding,''

\bibitem{Li2015}
X.~Li, C.~Zhang, W.~Li, Y.~A. Kuzovkina, and D.~Weiner, ``Who lives in greener
  neighborhoods? the distribution of street greenery and its association with
  residents' socioeconomic conditions in hartford, connecticut, usa,'' {\em
  Urban Forestry and Urban Greening}, vol.~14, pp.~751--759, 2015.

\bibitem{gebru}
T.~Gebru, J.~Krause, Y.~Wang, D.~Chen, J.~Deng, E.~L. Aiden, and L.~Fei-Fei,
  ``Using deep learning and google street view to estimate the demographic
  makeup of neighborhoods across the united states,'' {\em Proceedings of the
  National Academy of Sciences}, vol.~114, no.~50, pp.~13108--13113, 2017.

\bibitem{imagenet}
J.~Deng, W.~Dong, R.~Socher, L.-J. Li, K.~Li, and L.~Fei-Fei, ``Imagenet: A
  large-scale hierarchical image database,'' in {\em 2009 IEEE conference on
  computer vision and pattern recognition}, pp.~248--255, Ieee, 2009.

\bibitem{dropout}
N.~Srivastava, G.~E. Hinton, A.~Krizhevsky, I.~Sutskever, and R.~Salakhutdinov,
  ``Dropout: a simple way to prevent neural networks from overfitting.,'' {\em
  Journal of Machine Learning Research}, vol.~15, no.~1, pp.~1929--1958, 2014.

\bibitem{integrated-gradients}
M.~Sundararajan, A.~Taly, and Q.~Yan, ``Axiomatic attribution for deep
  networks,'' {\em CoRR}, vol.~abs/1703.01365, 2017.

\bibitem{kokhlikyan2020captum}
N.~Kokhlikyan, V.~Miglani, M.~Martin, E.~Wang, B.~Alsallakh, J.~Reynolds,
  A.~Melnikov, N.~Kliushkina, C.~Araya, S.~Yan, and O.~Reblitz-Richardson,
  ``Captum: A unified and generic model interpretability library for pytorch,''
  2020.

\bibitem{scikit-learn}
F.~Pedregosa, G.~Varoquaux, A.~Gramfort, V.~Michel, B.~Thirion, O.~Grisel,
  M.~Blondel, P.~Prettenhofer, R.~Weiss, V.~Dubourg, J.~Vanderplas, A.~Passos,
  D.~Cournapeau, M.~Brucher, M.~Perrot, and E.~Duchesnay, ``Scikit-learn:
  Machine learning in {P}ython,'' {\em Journal of Machine Learning Research},
  vol.~12, pp.~2825--2830, 2011.

\bibitem{pytorch}
A.~Paszke, S.~Gross, F.~Massa, A.~Lerer, J.~Bradbury, G.~Chanan, T.~Killeen,
  Z.~Lin, N.~Gimelshein, L.~Antiga, A.~Desmaison, A.~Kopf, E.~Yang, Z.~DeVito,
  M.~Raison, A.~Tejani, S.~Chilamkurthy, B.~Steiner, L.~Fang, J.~Bai, and
  S.~Chintala, ``Pytorch: An imperative style, high-performance deep learning
  library,'' in {\em Advances in Neural Information Processing Systems 32},
  pp.~8024--8035, Curran Associates, Inc., 2019.

\bibitem{adam}
D.~P. Kingma and J.~Ba, ``Adam: A method for stochastic optimization,'' in {\em
  ICLR (Poster)}, 2015.

\end{thebibliography}

\end{document}